\documentclass[11pt,a4paper]{article}
\usepackage[hyperref]{naaclhlt2018}
\usepackage{times}
\usepackage{latexsym}
\usepackage{graphics}

\usepackage{url}

\aclfinalcopy % Uncomment this line for the final submission

\usepackage{latexsym}
\usepackage{amsmath}
\usepackage{amssymb}
\usepackage{algorithm}
\usepackage{algpseudocode}
\usepackage{pifont}
\usepackage[font={footnotesize}]{caption}
\usepackage{sidecap}
\usepackage{subcaption}
\usepackage{graphicx}
\usepackage{booktabs}
\usepackage{float}
\usepackage{multirow}
\usepackage{url}           
\usepackage{booktabs}     
\usepackage{amsfonts}      
\usepackage{nicefrac}      
\usepackage{microtype}     
\usepackage{wrapfig}
\usepackage{graphicx}
\usepackage{amsmath}
\usepackage{makecell}
\usepackage[justification=justified]{caption}

\title{Visual Referring Expression Recognition: \\ What Do Systems Actually Learn?}
\author{
  Volkan Cirik, Louis-Philippe Morency, Taylor Berg-Kirkpatrick  \\
Carnegie Mellon University\\
\texttt{\{vcirik,morency,tberg\}@cs.cmu.edu} \\
}
\begin{document}
\maketitle
\begin{abstract}
We present an empirical analysis of the state-of-the-art systems for referring expression recognition -- the task of identifying the object in an image referred to by a natural language expression -- with the goal of gaining insight into how these systems reason about language and vision. Surprisingly, we find strong evidence that even sophisticated and linguistically-motivated models for this task may ignore the linguistic structure, instead relying on shallow correlations introduced by unintended biases in the data selection and annotation process. For example, we show that a system trained and tested on the input image \emph{without the input referring expression} can achieve a precision of 71.2\% in top-2 predictions.
Furthermore, a system that predicts only the \emph{object category} given the input can achieve a precision of 84.2\% in top-2 predictions. These surprisingly positive results for what should be deficient prediction scenarios suggest that careful analysis of what our models are learning -- and further, how our data is constructed -- is critical as we seek to make substantive progress on grounded language tasks.
\end{abstract}
\newcommand \figurethree{
\begin{figure*}[h!]
\vspace{-5pt}
\centering
\includegraphics[width=0.88\linewidth]{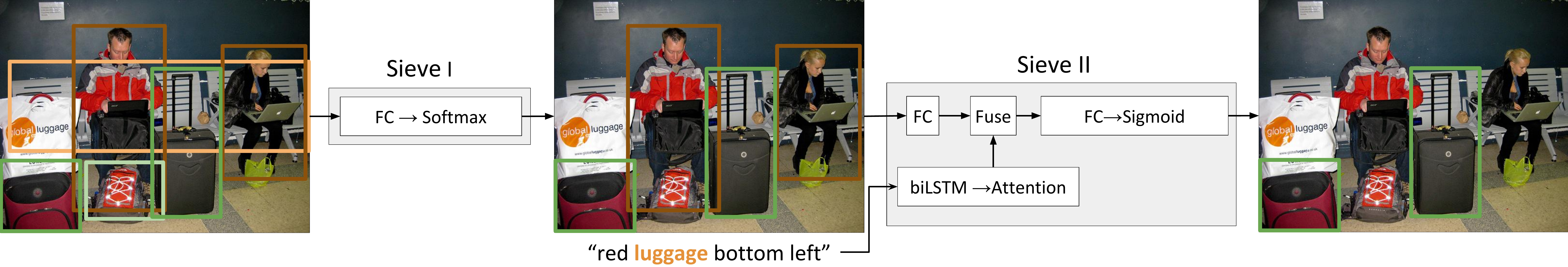}
\caption{Overview of Neural Sieves. Sieve I filters object types having multiple instances. Sieve II filters objects of one category mentioned in referring expression.
Objects of the same category have the same color frames. Best seen in color.}\label{fig2}
\vspace{-5pt}
\end{figure*}
}
\newcommand \tableone{
\begin{table*}[t]
\small
\centering
\begin{tabular}{@{}lrrrrrrr@{}}
\toprule
Model                  & \multicolumn{1}{l}{P@1} & \multicolumn{1}{l}{P@2} & \multicolumn{1}{l}{P@3} & \multicolumn{1}{l}{P@4} & \multicolumn{1}{l}{P@5} & \multicolumn{1}{l}{Object Relationships} & \multicolumn{1}{l}{Referring Expression}  \\ \midrule
CMN                    & .705 & .926 & .979 & .993 & .998 & $\checkmark$  &  $\checkmark$                \\
LSTM+CNN-MIL           & .684 & .907 & .972 & .993 & .997 & $\checkmark$  &  $\checkmark$                   \\
Neural Sieves (I + II + III)     & .694 & .912 & .975 & .991 & .991 & & $\checkmark$                   \\ \midrule
Neural Sieves (I)                & .401 & .712 & .866 & .935 & .967 & & \\
Neural Sieves (III)              & .691 & .916 & .974 & .991 & .996 & & $\checkmark$\\
Neural Sieves (I + III)          & .688 & .908 & .972 & .990 & .995  & & $\checkmark$                    \\
Neural Sieves (I + Oracle + III) & .723 & .947 & .991 & .998 & .999 & & $\checkmark$                     \\ \bottomrule
\end{tabular}
\caption{Results for baseline models and ablation studies for Neural Sieves. Neural Sieves achieve competitive results with a simple approach. Note that Sieve 1 is more accurate (71.2\%) than the state-of-the-art model (70.5\%) if we take top two predicted boxes without using the referring expression.}
\label{tab1}
\vspace{-5pt}
\end{table*}
}
\newcommand \tabletwo{
\begin{table*}[h]
%\small
\centering
\begin{tabular}{@{}lllll@{}}
\toprule
         & Shuffled & Noun \& Adjectives & Nouns & Adjectives \\ \midrule
CMN     & .675 (.030)     & .687 (.018)            & .642 (.063) & .585 (.120)      \\
LSTM+CNN-MIL & .630 (.054)    & .644 (.040)             & .597 (.087) & .533 (.151)      \\ \bottomrule
\end{tabular}
\caption{Precision@$1$ accuracies for the state-of-the-art models when the referring expression input is changed. The numbers in parentheses show the drop in accuracy compared to Table~\ref{tab1}. Models show an overly-stable behavior to the perturbations and still achieve good results.}
\label{tab2}
\vspace{-5pt}
\end{table*}
}
\newcommand \tablethree{
\begin{table*}[h]
\small
\centering
\begin{tabular}{@{}lllll@{}}
\toprule
Model                                & Metric & Score & Object Relationships & Referring Expression \\ \midrule
CMN                                  & p\@1    & .705  &  $\checkmark$                    &  $\checkmark$                    \\
LSTM+CNN-MIL                         & p\@1    & .684  &  $\checkmark$                    &  $\checkmark$                    \\
CMNloc                               & p\@1    & .691  &                      & $\checkmark$                     \\ \midrule
Neural Sieve I                       & p\@1    & .401  &                      &                      \\
Neural Sieve I                       & p\@2    & .712  &                      &                      \\
Neural Sieve I + II                  & p\@1    & .488  &                      &   $\checkmark$                   \\
Neural Sieve I + II                  & p\@2    & .842  &                      &   $\checkmark$                   \\
Neural Sieve I + II(Oracle)          & p\@1    & .577  &                      &   $\checkmark$                   \\
Neural Sieve I + II(Oracle)          & p\@2    & .905  &                      &    $\checkmark$                  \\
Neural Sieve I + CMNloc              & p\@1    & .688  &                      &   $\checkmark$                   \\
Neural Sieve I(Oracle) + CMNloc      & p\@1    & .708  &                      &    $\checkmark$                  \\
Neural Sieve I + II + CMNloc         & p\@1    & .694  &                      &   $\checkmark$                   \\
Neural Sieve I + II(Oracle) + CMNloc & p\@1    & .723  &                      & $\checkmark$                     \\ \bottomrule
\end{tabular}
\end{table*}
}
\newcommand \tablefour{
\begin{table*}[h]
\centering
\small
\begin{tabular}{@{}lrrrr@{}}
\toprule
\multirow{2}{*}{Model} & \multicolumn{4}{l}{Number of Boxes of the Same Object Type (Percentage of Data)} \\
                       & 2 (55\%)         & 3 (27.8 \%)         & 4 (17.2 \%)        & All (100\%)        \\ \midrule
CMN                    & .752             & .668                & .614               & .705               \\
CMNloc                 & .744             & .646                & .596               & .691               \\
CNN+LSTM-MIL           & .732             & .652                & .582               & .684               \\ \bottomrule
\end{tabular}
\end{table*}

}
\newcommand \tablefive{
\begin{table}[h]
\centering
\small
\begin{tabular}{@{}lrrr@{}}
\toprule
Model       & No Perturbation & Shuffled & $\Delta$ \\ \midrule
CMN         & .705 & .675     & -.030      \\
LSTM+CNN-MIL& .684 & .630     & -.054      \\ 
\bottomrule
\end{tabular}
\caption{Results for Shuffling Word Order for Referring Expressions. $\Delta$ is the difference between no perturbation and shuffled version of the same system.}\label{tab:syntax}
\vspace{-15pt}
\end{table}
}
\newcommand \tablesix{
\begin{table}[h]
\centering
\resizebox{\columnwidth}{!}{
\begin{tabular}{@{}lrrr@{}}
\toprule
Models       & Noun \& Adj ($\Delta$)& Noun  ($\Delta$)& Adj ($\Delta$)\\ \midrule
CMN          & .687 (-.018)              & .642 (-.063) & .585 (-.120)     \\
LSTM+CNN-MIL & .644 (-.040)              & .597 (-.087) & .533 (-.151)     \\ \bottomrule
\end{tabular}
}
\caption{Results with discarded word categories. Numbers in parentheses are $\Delta$, the difference between the best performing version of the original model.}\label{tab:lexical}
\vspace{-5pt}
\end{table}
}
\newcommand \tableseven{
\begin{table}[h]
\centering
\small
\begin{tabular}{@{}llllll@{}}
\toprule
Model & P@1  & P@2  & P@3  & P@4  & P@5  \\ \midrule
CMN          & .705 & .926 & .979 & .993 & .998 \\
CMN ``image-only'' & .411 & .731 & .885 & .948 & .977 \\
Random Baseline & .204 & .403 & .557 & .669 & .750\\
\bottomrule
\end{tabular}
\caption{Results with discarded referring expressions. Surprisingly, the top-2 prediction (73.1\%) of the ``image-only'' model is better than the top prediction of the state-of-the-art (70.5\%).}\label{tab:refexp}
\vspace{-5pt}
\end{table}
}
\newcommand \tableeight{
\begin{table}[t]
%\begin{SCtable}
\small
\centering
\begin{tabular}{@{}lll@{}}
\toprule
Model                        & precision@$k$        & Accuracy \\ \midrule
CMN    & 1                              & .705 \\
CMN    & 2                              & .926 \\
CMN    & 3                              & .979 \\
\midrule
LSTM+CNN-MIL  & 1                        & .684 \\
LSTM+CNN-MIL  & 2                        & .907 \\ 
LSTM+CNN-MIL  & 3                        & .972 \\          \midrule
Neural Sieve I & 1 & .401 \\
Neural Sieve I & 2 & .712 \\
Neural Sieve I & 3 & .866 \\
%Neural Sieve I & 4 & .935 & \\ 
\midrule
Neural Sieve I + II & 1 & .488 \\
Neural Sieve I + II &  2 & .842\\ 
Neural Sieve I + II & 3 & .953 \\ 
%Neural Sieve I + II & 4 & .992 & \\ 
\bottomrule
\end{tabular}
\caption{Precision@$k$ accuracies for Neural Sieves and state-of-the-art systems. Note that even without using the referring expression, Sieve I is able to reduce the number of candidate boxes to 3 for 86.6\% of the instances.
When we further predict the type of objects with Sieve II, the number of candidate boxes is reduced to 2 for 84.2\% of the instances.}\label{tab:sieve}
\vspace{-5pt}
\end{table}
%\end{SCtable}
}
\section{Introduction} 
\label{sec:introduction}
There has been increasing interest in modeling natural language in the context of a visual grounding. Several benchmark datasets have recently been introduced for describing a visual scene with natural language \cite{chen2015microsoft}, describing or localizing specific objects in a scene \cite{KazemzadehOrdonezMattenBergEMNLP14,mao2016generation}, answering natural language questions about the scenes \cite{antol2015vqa}, and performing visually grounded dialogue \cite{das2016visual}. Here, we focus on referring expression recognition (RER) -- the task of identifying the object in an image that is referred to by a natural language expression produced by a human \cite{KazemzadehOrdonezMattenBergEMNLP14,mao2016generation,hu2016natural,rohrbach2016grounding,yu2016modeling,nagaraja16refexp,hu2017modeling,cirik2018using}.

Recent work on RER has sought to make progress by introducing models that are better capable of reasoning about linguistic structure \cite{hu2017modeling,nagaraja16refexp} -- however, since most of the state-of-the-art systems involve complex neural parameterizations, what these models actually learn has been difficult to interpret.
This is concerning because several post-hoc analyses of related tasks \cite{zhou2015simple,devlin2015exploring,agrawal2016analyzing,jabri2016revisiting,goyal2016making} have revealed that some positive results are actually driven by superficial biases in datasets or shallow correlations without deeper visual or linguistic understanding. 
Evidently, it is hard to be completely sure if a model is performing well for the right reasons.

To increase our understanding of how RER systems function, we present several analyses inspired by approaches that probe systems with perturbed inputs \cite{jia2017adversarial} and employ simple models to exploit and reveal biases in datasets \cite{chen2016thorough}.
First, we investigate whether systems that were designed to incorporate linguistic structure actually require it and make use of it.
To test this, we perform perturbation experiments on the input referring expressions.
Surprisingly, we find that models are robust to shuffling the word order and limiting the word categories to nouns and adjectives.
Second, we attempt to reveal shallower correlations that systems might instead be leveraging to do well on this task. We build two simple systems called Neural Sieves: one that completely \textit{ignores} the input referring expression and another that only predicts the category of the referred object from the input expression.
Again, surprisingly, both sieves are able to identify the correct object with surprising precision in top-2 and top-3 predictions.
When these two simple systems are combined, the resulting system achieves precisions of 84.2\% and 95.3\% for top-2 and top-3 predictions, respectively.
These results suggest that to make meaningful progress on grounded language tasks, we need to pay careful attention to what and how our models are learning, and whether our datasets contain exploitable bias.
\vspace{-5pt}
\section{Related Work}\label{sec:related}
\vspace{-5pt}
Referring expression recognition and generation is a well studied problem in intelligent user interfaces \cite{chai2004probabilistic}, human-robot interaction \cite{fang2012integrating,chai2014collaborative, williams2016situated}, and situated dialogue \cite{kennington2017simple}.
\citet{KazemzadehOrdonezMattenBergEMNLP14} and \citet{mao2016generation} introduce two benchmark datasets for referring expression recognition.
Several models that leverage linguistic structure have been proposed. \citet{nagaraja16refexp} propose a model where the target and supporting objects (i.e.~objects that are mentioned in order to disambiguate the target object) are identified and scored jointly. The resulting model is able to localize supporting objects without direct supervision.
\citet{hu2017modeling} introduce a compositional approach for the RER task. They assume that the referring expression can be decomposed into a triplet consisting of the target object, the supporting object, and their spatial relationship. This structured model achieves state-of-the-art accuracy on the Google-Ref dataset. \citet{cirik2018using} propose a type of neural modular network \cite{andreas2016neural} where the computation graph is defined in terms of a constituency parse of the input referring expression. 

Previous studies on other tasks have found that the state-of-the-art systems may be successful for reasons different than originally assumed. For example, \citet{chen2016} show that a simple logistic regression baseline with carefully defined features can achieve competitive results for reading comprehension on CNN/Daily Mail datasets \cite{hermann2015teaching}, indicating that more sophisticated models may be learning relatively simple correlations.
Similarly, \citet{gururangan2018annotation} reveal bias in a dataset for semantic inference by demonstrating a simple model that achieves competitive results \textit{without looking at the premise}.
\figurethree
\vspace{-5pt}
\section{Analysis by Perturbation}\label{sec:analysis}
\vspace{-5pt}
In this section, we analyze how the state-of-the-art referring expression recognition systems utilize linguistic structure.
We conduct experiments with perturbed referring expressions where various aspects of the linguistic structure are obscured.
We perform three types of analyses: the first one studying syntactic structure (Section~\ref{ssec:syntax}), the second one focusing on the importance of word categories (Section~\ref{ssec:lexical}), and the final one analyzing potential biases in the dataset (Section~\ref{ssec:refexp}).

\subsection{Analysis Methodology}\label{ssec:analysis}%\vspace{-5pt}
To perform our analysis,
we take two state-of-the-art systems CNN+LSTM-MIL \cite{nagaraja16refexp} and CMN \cite{hu2017modeling} 
and train them from scratch with perturbed referring expressions.
We note that the perturbation experiments explained in next subsections are performed on all train and test instances.
All experiments are done on the standard train/test splits for the Google-Ref dataset \cite{mao2016generation}. Systems are evaluated using the precision@$k$ metric, the fraction of test instances for which the target object is contained in the model's top-$k$ predictions.
We provide further details of our experimental methodology in Section~\ref{sec:experiments}. 

\subsection{Syntactic Analysis by Shuffling Word Order}\label{ssec:syntax}%\vspace{-5pt}
In English, the word order is important for correctly understanding the syntactic structure of a sentence.
Both models we analyze use Recurrent Neural Networks (RNN) \cite{elman1990finding} with Long Short-Term Memory (LSTM) cells \cite{hochreiter1997long}.
Previous studies have shown that recurrent architectures can perform well on tasks where word order and syntax are important: for example, tagging \cite{lample2016neural}, parsing \cite{sutskever2014sequence}, and machine translation \cite{bahdanau2014neural}. We seek to determine whether recurrent models for RER depend on syntactic structure.

\noindent\textbf{Premise 1:} \textit{Shuffling the word order of an English referring expression will obscure its syntactic structure.}

\noindent We train CMN and CNN+LSTM-MIL with shuffled referring expressions as input and evaluate their performance.

\tablefive
Table~\ref{tab:syntax} shows accuracies for models with and without shuffled referring expressions.
The column with $\Delta$ shows the difference in accuracy compared to the best performing model without shuffling. 
The drop in accuracy is surprisingly low. 
Thus, we conclude that these models do not strongly depend on the syntactic structure of the input expression and may instead leverage other, shallower, correlations.

\subsection{Lexical Analysis by Discarding Words}\label{ssec:lexical}%\vspace{-5pt}
Following the analysis presented in Section~\ref{ssec:syntax}, we are curious to study what other aspects of the input referring expression may be essential for the state-of-the-art performance.
If the syntactic structure is largely unimportant, it may be that spatial relationships can be ignored.
Spatial relationships between objects are usually represented by prepositional phrases and verb phrases.
In contrast, simple descriptors (e.g.~green) and object types (e.g.~table) are most often represented by adjectives and nouns, respectively. By discarding all words in the input that are not nouns or adjectives, we hope to test whether spatial relationships are actually important to the state-of-the-art models.
Notably, both systems we test were specifically designed to model object relationships.

\noindent\textbf{Premise 2:} \textit{Keeping only nouns and adjectives from the input expression will obscure the relationships between objects that the referring expression describes.}

Table~\ref{tab:lexical} shows accuracies resulting from training and testing these models on only the nouns and adjectives in the input expression. Our first observation is that the accuracies of models drop the most when we discard the nouns (the rightmost column in Table~\ref{tab:lexical}).
\tablesix
This is reasonable since nouns define the types of the objects referred to in the expression. Without nouns, it is extremely difficult to identify which objects are being described.
Second, although both systems we analyze model the relationship between objects, discarding verbs and prepositions, which are essential in determining the relationship among objects, does not drastically reduce their performance (the second column in Table~\ref{tab:lexical}).
This may indicate the superior performance of these systems does not specifically come from their 
modeling approach for object relationships.

\subsection{Bias Analysis by Discarding Referring Expressions}\label{ssec:refexp}%\vspace{-5pt}
\citet{goyal2016making} show that some language and vision datasets have exploitable biases. 
Could there be a dataset bias that is exploited by the models for RER?
% The results in the previous sections hints that it might be the case.
% Here we take our analysis approach in the previous approach to the extreme and \textit{discard the whole referring expression} to observe the extent of possible visual bias.

\noindent\textbf{Premise 3:} \textit{ Discarding the referring expression entirely and keeping only the input image creates a deficient prediction problem: achieving high performance on this task indicates dataset bias.}

We train CMN by removing all referring expressions from train and test sets.
We call this model ``image-only'' since it ignores the referring expression and will only use the input image. 
We compare the CMN ``image-only'' model with the state-of-the-art configuration of CMN and a \textit{random baseline}. 
\tableseven
Table~\ref{tab:refexp} shows precision@$k$ results.
The ``image-only'' model
is able to surpass the random baseline by a large margin.
This result indicates that the dataset is biased, likely as a result of the data selection and annotation process.
During the construction of the dataset, \citet{mao2016generation} annotate an object box only if there are at least 2 to 4 objects of the same type in the image.
% to make sure that the object is ambiguous in the scene.
Thus, only a subset of object categories ever appears as targets because some object types rarely occur multiple times in an image.
In fact, out of 90 object categories in MSCOCO, 43 of the object categories are selected as the target objects less than 1\% of the time they occur in images.
% which results in a long-tail distribution of selected object categories.
%%
This potentially explains the relative high performance of the ``image-only'' system.

\subsection{Discussion}%\vspace{-5pt}
The previous analyses indicate that exploiting bias in the data selection process and leveraging shallow linguistic correlations with the input expression may go a long way towards achieving high performance on this dataset.
First, it may be possible to simplify the decision of picking an object to a much smaller set of candidates without even considering the referring expression. 
Second, because removing all words except for nouns and adjectives only marginally hurt performance for the systems tested, it may be possible to further reduce the set of candidates by focusing only on simple properties like the category of the target object rather than its relations with the environment or with adjacent objects.  
\vspace{-5pt}
\section{Neural Sieves} \label{sec:neuralsieves}
\vspace{-5pt}
We introduce a simple pipeline of neural networks, Neural Sieves, that attempt to reduce the set of candidate objects down to a much smaller set that still contains the target object given an image, a set of objects, and the referring expression describing one of the objects.

\paragraph{Sieve I: Filtering Unlikely Objects.}\label{sec:s1} 
Inspired by the results from Section~\ref{ssec:refexp}, we design an ``image-only'' model 
as the first sieve for filtering \textit{unlikely} objects.
For example in Figure~\ref{fig2}, Sieve I filters out the backpack and the bench from the list of bounding boxes since there is only one instance of these object types.
We use a similar parameterization of one of the baselines ($\text{CMN}_{LOC}$) proposed by \citet{hu2017modeling} for Sieve I and train it by only providing spatial and visual features for the boxes, ignoring the referring expression. More specifically, for visual features $r^{vis}$ of a bounding box of an object, we use Faster-RCNN \citep{ren2015faster}.
We use 5-dimensional vectors for spatial features $r^{spat} = [\frac{x_{min}}{W_V}, \frac{y_{min}}{H_V}, \frac{x_{max}}{W_V}, \frac{y_{max}}{W_V}, \frac{A_r}{A_V}  ]$ where $A_r$ is the size and $[x_{min}, y_{min}, x_{max}, y_{max}]$ are coordinates for bounding box $r$ and $A_V$, $W_V$, $H_V$ are the area, the width, and the height of the input image $V$.
These two representations are concatenated as $r^{vis,spat} = [r^{vis} r^{spat}]$ for a bounding box $r$.

We parameterize Sieve I with a list of bounding boxes $R$ as the input with a parameter set $\Theta_I$ as follows:
\vspace{-10pt}
\begin{align}
	s_{I} & = W_{I}^{score} r^{vis,spat} \label{eq-s1-1} \\ 
    f_{I}(R; \Theta_{I}) & = softmax(s_{I}) \label{eq-s1-2}
\end{align}
Each bounding box is scored using a matrix $W^{score}_{I}$.
Scores for all bounding boxes are then fed to softmax to get a probability distribution over boxes.
The learned parameter $\Theta_{I}$ is the scoring  matrix $W_I^{score}$.
\paragraph{Sieve II: Filtering Based on Objects Categories}\label{sec:s2} After filtering \textit{unlikely} objects based only on the image, the second step is to determine which object category to keep as a candidate for prediction, filtering out the other categories.
For instance, in Figure~\ref{fig2}, only instances of suitcases are left as candidates after determining which type of object the input expression is talking about.
To perform this step, Sieve II takes the list of object candidates from  Sieve I and keeps objects having the same object category as the referred object.
Unlike Sieve I, Sieve II uses the referring expression to filter bounding boxes of objects.
We again use the baseline model of $\text{CMN}_{\text{LOC}}$ from the previous work \citep{hu2017modeling} for the parametrization of Sieve II with a minor modification: instead of predicting the referred object, we make a binary decision for each box of whether the object in the box is the same category as the target object.

More specifically, we parameterize Sieve II as follows:
\vspace{-10pt}
\begin{align}
	\hat{r}^{vis,spat} & = W^{vis,spat}_{II} r^{vis,spat} \label{eq-s2-1} \\ 
	z_{II} & = \hat{r}^{vis,spat} \odot f_{att}(T) \label{eq-s2-2} \\
    \hat{z}_{II} & =  z_{II} / \mid\mid z_{II} \mid\mid_{2} \label{eq-s2-3} \\ 
    s_{II} & =  W_{II}^{score} \hat{z}_{s2}  \label{eq-s2-4} \\
    f_{II}(T,R; \Theta_{II}) & =  sigmoid(s_{II}) \label{eq-s2-5}
\end{align}
We encode the referring expression $T$  into an embedding with $f_{att}(T)$ which uses an attention mechanism \cite{bahdanau2014neural} on top of a 2-layer bidirectional LSTM \citep{schuster1997bidirectional}.

We project bounding box features $r^{vis,spat}$ to the same dimension as the embedding of referring expression (Eq~\ref{eq-s2-1}).
Text and box representations are element-wise multiplied to get $z_{II}$ as a joint representation of the text and bounding box (Eq~\ref{eq-s2-2}).
We L2-normalize to produce $\hat{z}_{II}$ (Eq~\ref{eq-s2-3},~\ref{eq-s2-4}).
Box scores $s_{II}$ are calculated with a linear projection of the joint representation (Eq~\ref{eq-s2-4}) and fed to the sigmoid function for a binary prediction for each box.
The learned parameters $\Theta_{II}$ are $W_{II}^{vis,spat}$,$W_{II}^{score}$, and parameters of the encoding module $f_{att}$.
\vspace{-5pt}
\subsection{Filtering Experiments}\label{sec:experiments}
\vspace{-5pt}
We are interested in determining how accurate these simple neural sieves can be. High accuracy here would give a possible explanation for the high performance of more complex models.
%\vspace{-7pt}
\tableeight
\paragraph{Dataset.} For our experiments, we use Google-Ref \citep{mao2016generation} which is one of the standard benchmarks for referring expression recognition. It consists of around 26K images with 104K annotations.
We use their Ground-Truth evaluation setup where the ground truth bounding box annotations from MSCOCO \citep{lin2014microsoft} are provided to the system as a part of the input.
We used the split provided by \citet{nagaraja16refexp} where splits have disjoint sets of images.
We use precision@$k$ for evaluating the performance of models.
\paragraph{Implementation Details.} 
To train our models, we used stochastic gradient descent for 6 epochs with an initial learning rate of 0.01 and multiplied by 0.4 after each epoch.
Word embeddings were initialized using GloVe \citep{pennington2014glove} and finetuned during training.
We extracted features for bounding boxes using the fc7 layer output of Faster-RCNN VGG-16 network \citep{ren2015faster} pre-trained on MSCOCO dataset \citep{lin2014microsoft}.
Hyperparameters such as hidden layer size of LSTM networks were picked based on the best validation score.
For perturbation experiments, we did not perform any grid search for hyperparameters. We used hyperparameters of the previously reported best performing model in the literature.
We released our code for public use\footnote{\href{https://github.com/volkancirik/neural-sieves-refexp}{https://github.com/volkancirik/neural-sieves-refexp}}.

\paragraph{Baseline Models.} We compare Neural Sieves to the state-of-the-art models from the literature.
\textbf{LSTM + CNN - MIL} \citet{nagaraja16refexp} score \textit{target object-context object} pairs using LSTMs for processing the referring expression and CNN features for bounding boxes. The pair with the highest score is predicted as the referred object.
They use Multi-Instance Learning for training the model.
\textbf{CMN} \citep{hu2017modeling} is a neural module network with a tuple of object-relationship-subject nodes.
The text encoding of tuples is calculated with a two-layer bi-directional LSTM and an attention mechanism \citep{bahdanau2014neural} over the referring expression.
\vspace{-5pt}
\subsection{Results}
\vspace{-5pt}
Table~\ref{tab:sieve} shows the precision scores.
The referred object is in the top-2 candidates selected by Sieve I 71.2\% of the time and in the top-3 predictions 86.6\% of the time. 
Combining both sieves into a pipeline,
these numbers further increase to 84.2\% for top-2 predictions and to 95.3\% for top-3 predictions.
Considering the simplicity of Neural Sieve approach, these are surprising results: two simple neural network systems, the first one ignoring the referring expression, the second predicting only object type, are able to reduce the number of candidate boxes down to 2 on 84.2\% of instances.
\vspace{-5pt}
\section{Conclusion} \label{sec:conclusion}
\vspace{-5pt}
We have analyzed two RER systems by variously perturbing aspects of the input referring expressions: shuffling, removing word categories, and finally, by removing the referring expression entirely.
Based on this analysis, we proposed a pipeline of simple neural sieves that captures many of the easy correlations in the standard dataset.
Our results suggest that careful analysis is important both while constructing new datasets and while constructing new models for grounded language tasks. The techniques used here may be applied more generally to other tasks to give better insight into what our models are learning and whether our datasets contain exploitable bias.
\bibliography{naaclhlt2018}

\begin{thebibliography}{}
\expandafter\ifx\csname natexlab\endcsname\relax\def\natexlab#1{#1}\fi

\bibitem[{Agrawal et~al.(2016)Agrawal, Batra, and
  Parikh}]{agrawal2016analyzing}
Aishwarya Agrawal, Dhruv Batra, and Devi Parikh. 2016.
\newblock \href{https://doi.org/10.18653/v1/D16-1203}{Analyzing the behavior of
  visual question answering models}.
\newblock In {\em Proceedings of the 2016 Conference on Empirical Methods in
  Natural Language Processing\/}. Association for Computational Linguistics,
  pages 1955--1960.
\newblock \url{https://doi.org/10.18653/v1/D16-1203}.

\bibitem[{Andreas et~al.(2016)Andreas, Rohrbach, Darrell, and
  Klein}]{andreas2016neural}
Jacob Andreas, Marcus Rohrbach, Trevor Darrell, and Dan Klein. 2016.
\newblock Neural module networks.
\newblock In {\em Proceedings of the IEEE Conference on Computer Vision and
  Pattern Recognition (CVPR)\/}. pages 39--48.

\bibitem[{Antol et~al.(2015)Antol, Agrawal, Lu, Mitchell, Batra,
  Lawrence~Zitnick, and Parikh}]{antol2015vqa}
Stanislaw Antol, Aishwarya Agrawal, Jiasen Lu, Margaret Mitchell, Dhruv Batra,
  C~Lawrence~Zitnick, and Devi Parikh. 2015.
\newblock Vqa: Visual question answering.
\newblock In {\em Proceedings of the IEEE International Conference on Computer
  Vision\/}. pages 2425--2433.

\bibitem[{Bahdanau et~al.(2014)Bahdanau, Cho, and Bengio}]{bahdanau2014neural}
Dzmitry Bahdanau, Kyunghyun Cho, and Yoshua Bengio. 2014.
\newblock Neural machine translation by jointly learning to align and
  translate.
\newblock {\em arXiv preprint arXiv:1409.0473\/} .

\bibitem[{Chai et~al.(2004)Chai, Hong, and Zhou}]{chai2004probabilistic}
Joyce~Y Chai, Pengyu Hong, and Michelle~X Zhou. 2004.
\newblock A probabilistic approach to reference resolution in multimodal user
  interfaces.
\newblock In {\em Proceedings of the 9th international conference on
  Intelligent user interfaces\/}. ACM, pages 70--77.

\bibitem[{Chai et~al.(2014)Chai, She, Fang, Ottarson, Littley, Liu, and
  Hanson}]{chai2014collaborative}
Joyce~Y Chai, Lanbo She, Rui Fang, Spencer Ottarson, Cody Littley, Changsong
  Liu, and Kenneth Hanson. 2014.
\newblock Collaborative effort towards common ground in situated human-robot
  dialogue.
\newblock In {\em Proceedings of the 2014 ACM/IEEE international conference on
  Human-robot interaction\/}. ACM, pages 33--40.

\bibitem[{Chen et~al.(2016{\natexlab{a}})Chen, Bolton, and
  Manning}]{chen2016thorough}
Danqi Chen, Jason Bolton, and Christopher~D. Manning. 2016{\natexlab{a}}.
\newblock \href{https://doi.org/10.18653/v1/P16-1223}{A thorough examination of
  the cnn/daily mail reading comprehension task}.
\newblock In {\em Proceedings of the 54th Annual Meeting of the Association for
  Computational Linguistics (Volume 1: Long Papers)\/}. Association for
  Computational Linguistics, pages 2358--2367.
\newblock \url{https://doi.org/10.18653/v1/P16-1223}.

\bibitem[{Chen et~al.(2016{\natexlab{b}})Chen, Bolton, and Manning}]{chen2016}
Danqi Chen, Jason Bolton, and Christopher~D. Manning. 2016{\natexlab{b}}.
\newblock \href{http://www.aclweb.org/anthology/P16-1223}{A thorough
  examination of the cnn/daily mail reading comprehension task}.
\newblock In {\em Proceedings of the 54th Annual Meeting of the Association for
  Computational Linguistics (Volume 1: Long Papers)\/}. Association for
  Computational Linguistics, Berlin, Germany, pages 2358--2367.
\newblock \url{http://www.aclweb.org/anthology/P16-1223}.

\bibitem[{Chen et~al.(2015)Chen, Fang, Lin, Vedantam, Gupta, Doll{\'a}r, and
  Zitnick}]{chen2015microsoft}
Xinlei Chen, Hao Fang, Tsung-Yi Lin, Ramakrishna Vedantam, Saurabh Gupta, Piotr
  Doll{\'a}r, and C~Lawrence Zitnick. 2015.
\newblock Microsoft coco captions: Data collection and evaluation server.
\newblock {\em arXiv preprint arXiv:1504.00325\/} .

\bibitem[{Cirik et~al.(2018)Cirik, Berg-Kirkpatrick, and
  Morency}]{cirik2018using}
Volkan Cirik, Taylor Berg-Kirkpatrick, and Louis-Phillippe Morency. 2018.
\newblock Using syntax to ground referring expressions in natural images.
\newblock In {\em 32nd AAAI Conference on Artificial Intelligence (AAAI-18)\/}.

\bibitem[{Das et~al.(2016)Das, Kottur, Gupta, Singh, Yadav, Moura, Parikh, and
  Batra}]{das2016visual}
Abhishek Das, Satwik Kottur, Khushi Gupta, Avi Singh, Deshraj Yadav,
  Jos{\'e}~MF Moura, Devi Parikh, and Dhruv Batra. 2016.
\newblock Visual dialog.
\newblock {\em arXiv preprint arXiv:1611.08669\/} .

\bibitem[{Devlin et~al.(2015)Devlin, Gupta, Girshick, Mitchell, and
  Zitnick}]{devlin2015exploring}
Jacob Devlin, Saurabh Gupta, Ross Girshick, Margaret Mitchell, and C~Lawrence
  Zitnick. 2015.
\newblock Exploring nearest neighbor approaches for image captioning.
\newblock {\em arXiv preprint arXiv:1505.04467\/} .

\bibitem[{Elman(1990)}]{elman1990finding}
Jeffrey~L Elman. 1990.
\newblock Finding structure in time.
\newblock {\em Cognitive science\/} 14(2):179--211.

\bibitem[{Fang et~al.(2012)Fang, Liu, and Chai}]{fang2012integrating}
Rui Fang, Changsong Liu, and Joyce~Yue Chai. 2012.
\newblock Integrating word acquisition and referential grounding towards
  physical world interaction.
\newblock In {\em Proceedings of the 14th ACM international conference on
  Multimodal interaction\/}. ACM, pages 109--116.

\bibitem[{Goyal et~al.(2016)Goyal, Khot, Summers-Stay, Batra, and
  Parikh}]{goyal2016making}
Yash Goyal, Tejas Khot, Douglas Summers-Stay, Dhruv Batra, and Devi Parikh.
  2016.
\newblock Making the v in vqa matter: Elevating the role of image understanding
  in visual question answering.
\newblock {\em arXiv preprint arXiv:1612.00837\/} .

\bibitem[{Gururangan et~al.(2018)Gururangan, Swayamdipta, Levy, Schwartz,
  Bowman, and Smith}]{gururangan2018annotation}
Suchin Gururangan, Swabha Swayamdipta, Omer Levy, Roy Schwartz, Samuel~R
  Bowman, and Noah~A Smith. 2018.
\newblock Annotation artifacts in natural language inference data.
\newblock In {\em Proceedings of the 2018 Conference of the North American
  Chapter of the Association for Computational Linguistics: Human Language
  Technologies\/}. Association for Computational Linguistics.

\bibitem[{Hermann et~al.(2015)Hermann, Kocisky, Grefenstette, Espeholt, Kay,
  Suleyman, and Blunsom}]{hermann2015teaching}
Karl~Moritz Hermann, Tomas Kocisky, Edward Grefenstette, Lasse Espeholt, Will
  Kay, Mustafa Suleyman, and Phil Blunsom. 2015.
\newblock Teaching machines to read and comprehend.
\newblock In {\em Advances in Neural Information Processing Systems\/}. pages
  1693--1701.

\bibitem[{Hochreiter and Schmidhuber(1997)}]{hochreiter1997long}
Sepp Hochreiter and J{\"u}rgen Schmidhuber. 1997.
\newblock Long short-term memory.
\newblock {\em Neural computation\/} 9(8):1735--1780.

\bibitem[{Hu et~al.(2017)Hu, Rohrbach, Andreas, Darrell, and
  Saenko}]{hu2017modeling}
Ronghang Hu, Marcus Rohrbach, Jacob Andreas, Trevor Darrell, and Kate Saenko.
  2017.
\newblock Modeling relationships in referential expressions with compositional
  modular networks .

\bibitem[{Hu et~al.(2016)Hu, Xu, Rohrbach, Feng, Saenko, and
  Darrell}]{hu2016natural}
Ronghang Hu, Huazhe Xu, Marcus Rohrbach, Jiashi Feng, Kate Saenko, and Trevor
  Darrell. 2016.
\newblock Natural language object retrieval.
\newblock In {\em Proceedings of the IEEE Conference on Computer Vision and
  Pattern Recognition\/}. pages 4555--4564.

\bibitem[{Jabri et~al.(2016)Jabri, Joulin, and van~der
  Maaten}]{jabri2016revisiting}
Allan Jabri, Armand Joulin, and Laurens van~der Maaten. 2016.
\newblock Revisiting visual question answering baselines.
\newblock In {\em European conference on computer vision\/}. Springer, pages
  727--739.

\bibitem[{Jia and Liang(2017)}]{jia2017adversarial}
Robin Jia and Percy Liang. 2017.
\newblock Adversarial examples for evaluating reading comprehension systems.
\newblock {\em arXiv preprint arXiv:1707.07328\/} .

\bibitem[{Kazemzadeh et~al.(2014)Kazemzadeh, Ordonez, Matten, and
  Berg}]{KazemzadehOrdonezMattenBergEMNLP14}
Sahar Kazemzadeh, Vicente Ordonez, Mark Matten, and Tamara~L. Berg. 2014.
\newblock Referit game: Referring to objects in photographs of natural scenes.
\newblock In {\em EMNLP\/}.

\bibitem[{Kennington and Schlangen(2017)}]{kennington2017simple}
Casey Kennington and David Schlangen. 2017.
\newblock A simple generative model of incremental reference resolution for
  situated dialogue.
\newblock {\em Computer Speech \& Language\/} 41:43--67.

\bibitem[{Lample et~al.(2016)Lample, Ballesteros, Subramanian, Kawakami, and
  Dyer}]{lample2016neural}
Guillaume Lample, Miguel Ballesteros, Sandeep Subramanian, Kazuya Kawakami, and
  Chris Dyer. 2016.
\newblock Neural architectures for named entity recognition.
\newblock {\em arXiv preprint arXiv:1603.01360\/} .

\bibitem[{Lin et~al.(2014)Lin, Maire, Belongie, Hays, Perona, Ramanan,
  Doll{\'a}r, and Zitnick}]{lin2014microsoft}
Tsung-Yi Lin, Michael Maire, Serge Belongie, James Hays, Pietro Perona, Deva
  Ramanan, Piotr Doll{\'a}r, and C~Lawrence Zitnick. 2014.
\newblock Microsoft coco: Common objects in context.
\newblock In {\em European Conference on Computer Vision\/}. Springer, pages
  740--755.

\bibitem[{Mao et~al.(2016)Mao, Huang, Toshev, Camburu, Yuille, and
  Murphy}]{mao2016generation}
Junhua Mao, Jonathan Huang, Alexander Toshev, Oana Camburu, Alan~L Yuille, and
  Kevin Murphy. 2016.
\newblock Generation and comprehension of unambiguous object descriptions.
\newblock In {\em Proceedings of the IEEE Conference on Computer Vision and
  Pattern Recognition (CVPR)\/}. pages 11--20.

\bibitem[{Nagaraja et~al.(2016)Nagaraja, Morariu, and Davis}]{nagaraja16refexp}
Varun Nagaraja, Vlad Morariu, and Larry Davis. 2016.
\newblock Modeling context between objects for referring expression
  understanding.
\newblock In {\em ECCV\/}.

\bibitem[{Pennington et~al.(2014)Pennington, Socher, and
  Manning}]{pennington2014glove}
Jeffrey Pennington, Richard Socher, and Christopher~D Manning. 2014.
\newblock Glove: Global vectors for word representation.
\newblock In {\em EMNLP\/}. volume~14, pages 1532--1543.

\bibitem[{Ren et~al.(2015)Ren, He, Girshick, and Sun}]{ren2015faster}
Shaoqing Ren, Kaiming He, Ross Girshick, and Jian Sun. 2015.
\newblock Faster r-cnn: Towards real-time object detection with region proposal
  networks.
\newblock In {\em Advances in neural information processing systems\/}. pages
  91--99.

\bibitem[{Rohrbach et~al.(2016)Rohrbach, Rohrbach, Hu, Darrell, and
  Schiele}]{rohrbach2016grounding}
Anna Rohrbach, Marcus Rohrbach, Ronghang Hu, Trevor Darrell, and Bernt Schiele.
  2016.
\newblock Grounding of textual phrases in images by reconstruction.
\newblock In {\em European Conference on Computer Vision\/}. Springer, pages
  817--834.

\bibitem[{Schuster and Paliwal(1997)}]{schuster1997bidirectional}
Mike Schuster and Kuldip~K Paliwal. 1997.
\newblock Bidirectional recurrent neural networks.
\newblock {\em IEEE Transactions on Signal Processing\/} 45(11):2673--2681.

\bibitem[{Sutskever et~al.(2014)Sutskever, Vinyals, and
  Le}]{sutskever2014sequence}
Ilya Sutskever, Oriol Vinyals, and Quoc~V Le. 2014.
\newblock Sequence to sequence learning with neural networks.
\newblock In {\em Advances in neural information processing systems\/}. pages
  3104--3112.

\bibitem[{Williams et~al.(2016)Williams, Acharya, Schreitter, and
  Scheutz}]{williams2016situated}
Tom Williams, Saurav Acharya, Stephanie Schreitter, and Matthias Scheutz. 2016.
\newblock Situated open world reference resolution for human-robot dialogue.
\newblock In {\em The Eleventh ACM/IEEE International Conference on Human Robot
  Interaction\/}. IEEE Press, pages 311--318.

\bibitem[{Yu et~al.(2016)Yu, Poirson, Yang, Berg, and Berg}]{yu2016modeling}
Licheng Yu, Patrick Poirson, Shan Yang, Alexander~C Berg, and Tamara~L Berg.
  2016.
\newblock Modeling context in referring expressions.
\newblock In {\em European Conference on Computer Vision\/}. Springer, pages
  69--85.

\bibitem[{Zhou et~al.(2015)Zhou, Tian, Sukhbaatar, Szlam, and
  Fergus}]{zhou2015simple}
Bolei Zhou, Yuandong Tian, Sainbayar Sukhbaatar, Arthur Szlam, and Rob Fergus.
  2015.
\newblock Simple baseline for visual question answering.
\newblock {\em arXiv preprint arXiv:1512.02167\/} .

\end{thebibliography}
\bibliographystyle{acl_natbib}
\end{document}